\documentclass{Interspeech2024}

\interspeechcameraready

\title{Investigating the Effects of Large-Scale Pseudo-Stereo Data and Different Speech Foundation Model on Dialogue Generative Spoken Language Model}

\name[affiliation={1,2}]{Yu-Kuan}{Fu}
\name[affiliation={1}]{Cheng-Kuang}{Lee}
\name[affiliation={2}]{Hsiu-Hsuan}{Wang}
\name[affiliation={2}]{Hung-yi}{Lee}
\address{
  $^1$NVIDIA\\
  $^2$National Taiwan University
}
\email{r11942083@ntu.edu.tw, ckl@nvidia.com, b09902033@ntu.edu.tw, tlkagkb93901106@gmail.com}

\usepackage{algorithm}
\usepackage{algpseudocode}
\usepackage{hyperref}
\usepackage{multirow}
\usepackage{adjustbox,lipsum}

\keywords{spoken dialogue modeling, speech generation, human-computer interaction}

\begin{document}

\maketitle

\begin{abstract}
Recent efforts in Spoken Dialogue Modeling aim to synthesize spoken dialogue without the need for direct transcription, thereby preserving the wealth of non-textual information inherent in speech. However, this approach faces a challenge when speakers talk simultaneously, requiring stereo dialogue data with speakers recorded on separate channels—a notably scarce resource. To address this, we have developed an innovative pipeline capable of transforming single-channel dialogue data into pseudo-stereo data. This expanded our training dataset from a mere 2,000 to an impressive 17,600 hours, significantly enriching the diversity and quality of the training examples available. The inclusion of this pseudo-stereo data has proven to be effective in improving the performance of spoken dialogue language models. Additionally, we explored the use of discrete units of different speech foundation models for spoken dialogue generation.

\end{abstract}

\section{Introduction}
% Spoken dialogue is the most fundamental method for human
% to communicate with each other, characterized by spontaneous
% turn transitions and occasional intervals of silence or overlapping speech between turns. Within spoken dialogue, these moments of silence and overlapping speech, along with various non-verbal signals like laughter and sighs, or what can be termed as turn-taking behaviors, are natural to humans\cite{nguyen2022systematic}, and carry significant cues for understanding the conversation\cite{yngve1970getting,stivers2009universals}.

% However, the majority of current spoken dialogue systems (e.g., Siri) do not account for these signals. They typically follow a traditional pipeline: transcribing incoming speech into text, generating a textual response, and then converting it back into speech using a Text-to-Speech (TTS) model. This conventional approach falls short in capturing the richness and natural flow of human conversation. Specifically, it struggles to recreate overlapping speech and non-verbal vocalizations, resulting in unnatural dialogues.

Spoken dialogue, characterized by spontaneous turn transitions and occasional overlaps, embodies the natural flow of human communication \cite{nguyen2022systematic}. These moments of silence, laughter, and overlapping speech are vital cues within conversations \cite{schegloff1982discourse,yngve1970getting,stivers2009universals}. Yet, current dialogue systems like Siri often overlook these signals, relying on a conventional pipeline of transcribing, generating textual response, and converting it to speech, resulting in unnatural dialogues \cite{huang2023audiogpt}.

Advancements in self-supervised (SSL) models \cite{baevski2020wav2vec,hsu2021hubert,chen2022wavlm,oord2018representation} and textless spoken language modeling \cite{lakhotia2021generative,kharitonov2021text,borsos2023audiolm} now enable encoding speech signals directly into discrete tokens without implicitly transcribing to text. This preserves both verbal and nonverbal cues, aligning with human turn-taking behaviors. The introduction of the dialogue generative spoken language model (dGSLM) \cite{nguyen2023generative} marks a milestone, employing a dual tower transformer to process input tokens from separate channels and enhance dialogue generation. While dGSLM has shown to generate natural human dialogue, it faces challenges in generating semantic coherence dialogue due to limited data.
The training data of dGSLM is a notably scarce resource because it requires stereo dialogue data. 

Acquiring stereo dialogue data is challenging. In contrast, single-channel dialogue content is abundant. For example, in this paper, we gathered approximately 20k hours of podcasts. While the speakers of podcasts are mixed into one channel, which dGSLM can not directly use, we proposed a pipeline to automatically generate pseudo-stereo data through the following process: first, employing speaker diarization to pinpoint segments featuring two speakers; next, leveraging source separation techniques to isolate overlapping frames; and finally, applying speaker verification to allocate the separated speech to its respective speakers. This meticulous process yielded a substantial 15.6k hours of pseudo-stereo dialogue training data\footnote{We open source our dataset: \url{https://huggingface.co/datasets/YuKuanFu/Podcast_Dialogue}}.

Additionally, we explored the use of discrete units from state-of-the-art (SOTA) foundation models for dGSLM. Our investigation revealed that merely scaling foundation models led to poor vocoder performance when resynthesizing speech from discrete units. However, employing an ASR fine-tuned foundation model showed significant improvements across all aspects. By integrating the ASR fine-tuned model with our pseudo-stereo data, dGSLM excelled in producing dialogue semantic coherence\footnote{\label{fn:demo}Generation samples can be found at \url{https://anonymous78264.github.io/pseudo-stereo-data/?fbclid=IwAR0MGdFnQeUcnojhQGGk0HaAYgBxhnhblIpU3xnGRNfPPO_hxHOf6Ea_PGM}}.

% This paper has the following contributions:
% \begin{itemize}
% \item To address the scarcity of stereo dialogue data, we investigate the utilization of pseudo-stereo dialogue data derived from single-channel sources. This approach significantly expands the resources available for training dGSLM.
% \item We have conducted explorations into the enhancement of dGSLM by employing various discrete units sourced from foundational speech models.
% \end{itemize}
% \section{Introduction}

% Templates are provided on the conference website for Microsoft Word\textregistered, and \LaTeX. We strongly recommend \LaTeX\xspace
% which can be used conveniently in a web browser on \url{overleaf.com} where this template is available in the Template Gallery.
\section{Related Work}

\subsection{Speech Unit Language Modeling}
\label{rw:ulm}
The Textless NLP series of works introduced a framework to address speech tasks using discrete NLP approaches. This framework can be broadly divided into three components \cite{lakhotia2021generative,borsos2023audiolm}: encoding speech into discrete units by clustering its pretrained self-supervised learning (SSL) representations \cite{hsu2021hubert, chen2022wavlm, baevski2020wav2vec,oord2018representation}, autoregressively generating discrete units, and resynthesizing speech from these discrete units\cite{polyak2021speech}.

Quantized unit sequences are shorter than the original signal, significantly reducing computational costs and enabling the modeling of speech through NLP approaches. Additionally, this framework has successfully generated natural speech without textual information \cite{lakhotia2021generative}. With more fine-grained discrete units, it can preserve non-verbal vocalizations and even control the prosody of speech \cite{kharitonov2021text}.

% Textless NLP\footnote{\href{https://speechbot.github.io/}{Textless NLP project page: https://speechbot.github.io/}} series of works proposed a framework to solve speech tasks with discrete NLP approaches. This framework can be largely divided into trhee components: encode speech, a continuous signal, into discrete unit by clustering its pretrained self-supervised learning(SSL) representations(\cite{hsu2021hubert, chen2022wavlm, baevski2020wav2vec}), autoregressively generate discrete unit, and resynthesize speech from the discrete unit.

% Quantized unit sequences are shorter than the original signal, largely decrease the computational cost, and enable us to modeling speech via NLP approach. Moreover, this framework successfully generate natural speech without textual information \cite{lakhotia2021generative}, and with more fine-grained discrete unit, it can preserve more non-verbal vocalization, and even control the prosody of speech \cite{kharitonov2021text}.

\subsection{Spoken Dialogue Generation}
Current commercial spoken dialogue generation models (e.g., Siri, Alexa, Google Assistant) divide spoken dialogue into three components: ASR, text-base dialogue language model, and TTS \cite{huang2023audiogpt}. These models primarily focus on semantic content, often neglecting  other speech related information, resulting in unnatural dialogue.
\begin{figure}[t]
    \centering
    \includegraphics[width=1.0\linewidth]{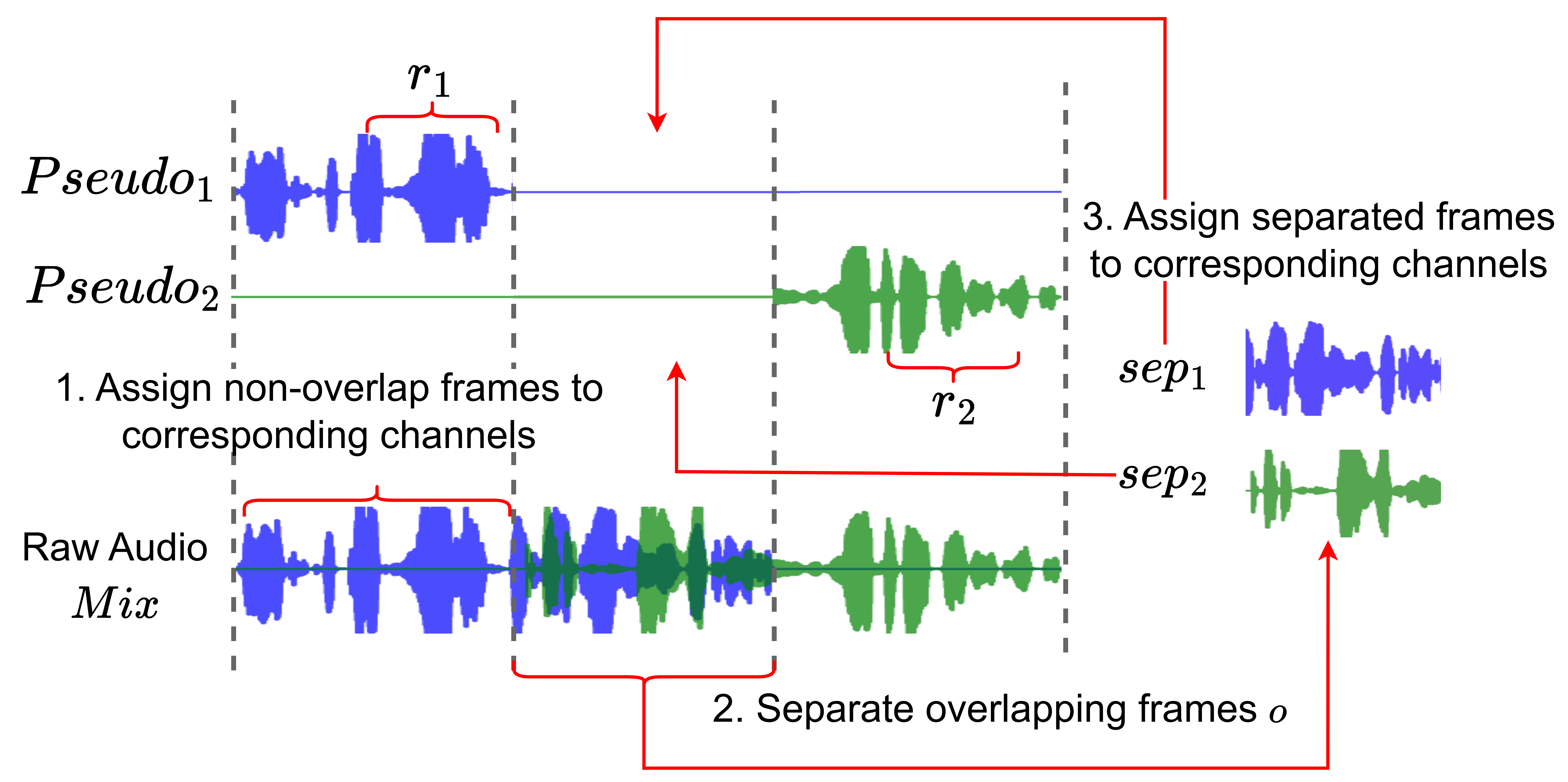}
    \caption{
        The pipeline of generating pseudo-stereo data from single-channel dialogue data. We split the process into 3 steps: speaker diarization, source separation, and speaker verification.
    }
    \label{fig:separation}
\end{figure}

Recent work has focused on generating a diverse set of spoken dialogues that incorporate turn-taking and acoustic information to mimic human-like conversational flow \cite{jin2022response, inoue2024realtime}. dGSLM \cite{nguyen2023generative} is the first spoken dialogue generation model that can successfully generated natural dialogue with non-verbal signals. It follows the three components mentioned in Sec.\ref{rw:ulm}. It first encodes two speakers by clustering the representations of SSL models separately, and generates two parallel discrete unit sequences from a dual-tower transformer language model. Moreover, it splits the next unit prediction into edge-unit prediction and duration prediction during training for better performance. Although dGSLM can generate natural dialogue, the lack of data results in poor performance in semantic coherence within dialogues.

\section{Method}
Our methodology is an enhancement of the dGSLM framework~\cite{nguyen2023generative}. 
dGSLM decomposes spoken dialogue generation into three components: the Speech-to-Units encoder, the dual-tower unit dialogue language model, and the Units-to-Speech vocoder. The original dGSLM model encountered difficulties in producing semantically coherent dialogue, a problem attributed to insufficient data and suboptimal unit encoding~\cite{nguyen2023generative}.
Thus in this study, we places emphasis on two main objectives: augmenting the training of dGSLM with additional data and improving the performance of unit encoding. 

The following sections present a detailed overview of our approach, covering the process of generating pseudo-stereo data from single-channel recording and the Speech-to-Units encoders used.

%\subsection{Data Preparation}

\subsection{Pseudo-stereo Data Generation}
\label{sec:pipeline}
dGSLM utilized the Fisher Telephone conversation collection protocol \cite{cieri2004fisher} as the training dataset for its components. The Fisher dataset is a well-known benchmark for spoken dialogue datasets, containing around 2000 hours of telephonic dialogues. Each conversation in the Fisher dataset involves two speakers recorded in separate audio channels, resulting in a stereo audio setup.

However, most available dialogue data do not separate speakers, often existing as single-channel audio. To overcome this limitation, we developed a pipeline to disentangle speakers from single-channel audio and store their speech into separate channels, creating pseudo-stereo data. 
%This approach expanded the training dataset for dGSLM, providing a more diverse dataset for improved model learning and performance. 
This strategy enhanced the training dataset for dGSLM by incorporating a wider variety of data, which enriched the model's learning process and boosted its overall performance.
The framework of this approach is illustrated in Fig.~\ref{fig:separation}. 
This process is divided into three steps:

\subsubsection{Speaker Diarization}
We used a Speaker Diarization model ($SD$) \cite{bredin2023pyannote,park2022multi} to disentangle speakers from a segment of raw audio ($Mix$), resulting in a set of speaker-duration tuples:
\begin{equation}
SD(Mix) = \{ (p, s, e) \mid p \in \{p_1, p_2\},\, e > s \}
\end{equation}
% \begin{equation}
% \cup_{i,j}\{(p_1, s_i, e_i)\}\cup\{(p_2, s_j, e_j)\}=SD(Mix)
% \end{equation}
Here, we only preserved segments that contains exactly two speakers. $p_1$ and $p_2$ represent the first and second speakers, while $s$ and $e$ are the start and end frames of the segment that $p$ is speaking.

We created a 2-channel audio ($Psuedo$) filled with silences, matching the length of $Mix$ (we use $Psuedo_1$, $Psuedo_2$ to indicate its first, second channel respectively). With the output of $SD$, we identified non-overlapping frames ($O'$) and overlapping frames ($O$):
\begin{equation}
\begin{aligned}
    O' &= \{(p, s, e) \mid \text{only p is speaking from } s \text{ to }e\} \\
    O &=\{(s,e) \mid p_1, p_2 \text{ are both speaking from } s \text{ to } e \}
\end{aligned}
\end{equation}

We then assigned non-overlapping intervals in $Mix$ to $Pseudo$ accoring to $O'$ (correspond to the step 1 in Fig. \ref{fig:separation}):
\begin{algorithmic}[1]
    \For{$i$ in 1, 2;}
        \For{$(p_i, s, e)$ in $O'$;}
            \State $Pseudo_{i}[s:e] \gets \text{Mix}[s:e]$
        \EndFor
    \EndFor
\end{algorithmic}

\subsubsection{Source Separation}
 Source separation model ($SS$) \cite{subakan2021attention,luo2020dual,Luo_2019} can effectively separate the overlapping speech ($o$):
\begin{equation}
    SS(o) = sep_1, sep_2
\end{equation}
Here, $sep_1$ and $sep_2$ represent the separated speech for the two speakers (correspond to Step 2 Fig. \ref{fig:separation}). 
%However, at this stage, we do not yet know which speaker corresponds to $p1$ and $p2$.
However, at this juncture, it remains unclear which of these segments can be attributed to speaker $p_1$ and which to speaker $p_2$.

Note that we can not directly apply $SS$ on entire $Mix$, as current source separation algorithms struggle with sparse overlapping speech \cite{cosentino2020librimix}.

\subsubsection{Speaker Verification}
We match $sep_1$ and $sep_2$ their corresponding speaker by speaker verification ($SV$) \cite{desplanques2020ecapa,koluguri2022titanet}. $SV$ compares two audio clips and returns the similarity of their speaker embeddings (correspond to the step 3 in Fig.\ref{fig:separation}):
\begin{equation}
    SV(r_i, sep_j) = Sim_{i,j}
\end{equation}
Here, $r_i$ is a reference clip randomly selected from non-overlapping frames from $p_i$ (see Fig.\ref{fig:separation}). We assigned $sep_1$ and $sep_2$ based on the similarities:

\begin{algorithmic}[1]
\For{$(s, e)$ in $O$;}
    \State $sep_1, sep_2 = SS(Mix[s: e])$
    \State $Sim_{i, j} = SV(r_i, sep_j)$ for $i$, $j$ in $(1, 2) \times (1,2)$
    \If{$Sim_{1, 1} + Sim_{2, 2} > Sim_{1, 2} + Sim_{2, 1}$}
        \State $Pseudo_1[s: e], Pseudo_2[s: e] \gets sep_1, sep_2$
    \Else
        \State $Pseudo_1[s: e], Pseudo_2[s: e] \gets sep_2, sep_1$
    \EndIf
\EndFor
\end{algorithmic}

By employing this pipeline, we successfully disentangled speakers in spoken dialogue, enabling scalable creation of diverse and large dialogue audio corpora.

\begin{table*}[]
    \caption{This table shows the turn-taking metrics. We report their difference with ground truth. Dur. indicates duration of events (seconds), and Occur. indicates occurence (times)}
    \centering
    \vspace{1pt}
    \begin{adjustbox}{width=\columnwidth*2,center}
    \begin{tabular} {l c c|@{\extracolsep{4pt}}cc cc cc cc@{}}
        \toprule
         & \multirow{2}{*}{Encoder} & \multirow{2}{*}{Type} & \multicolumn{2}{c}{$\Delta$ IPU} &
        \multicolumn{2}{c}{$\Delta$ Gap} & \multicolumn{2}{c}{$\Delta$ Overlap} & \multicolumn{2}{c}{$\Delta$ Pause} \\
        \cline{4-5}
        \cline{6-7}
        \cline{8-9}
        \cline{10-11}

        %\cmidrule(l{2pt}r{2pt}){4-6}  \cmidrule(l{2pt}r{2pt}){7-9} \cmidrule(l{2pt}r{2pt}){10-12}
         &  &  & Dur. & Occur.
        & Dur. & Occur. 
        & Dur. & Occur.
        & Dur. & Occur. \\
        \midrule
        
        \textit{(a)} & Ground Truth &
        & 0 (56.86) & 0 (19.86)
        & 0 (2.61) & 0 (2.88)
        & 0 (4.29) & 0 (3.96)
        & 0 (4.83) & 0 (7.42)\\
        \midrule
        \multicolumn{10}{l}{\textbf{\textsc{Fisher / Fisher + Pseudo-stereo Aduio}}} \\
        \textit{(b)} & \multirow{2}{*}{WavLM} & base+
        & 3.17 / -3.96 & 3.39 / -1.96
        & -0.37 / 1.02 & 1.49 / -0.57
        & 2.23 / -0.83 & 4.47 / 0.95
        & -0.57 / 2.1
        & 0.74 / 1.92 \\
        \textit{(c)} &  & large
        & -3.72 / 5.36 & 9.96 / 2.93 
        & 0.43 / -1.00 & 3.58 / 0.91
        & 0.22 / 3.14 & 3.72 / 4.47
        & 3.5 / -1.24 & 5.54 / 0.62 \\
        \midrule
        \textit{(d)} & \multirow{3}{*}{HuBERT} & base
        & 4.35 / 9.38 & 5.66 / 5.5
        & -0.35 / -0.94  & 2.06 / 0.91
        & 2.64 / 6.25 & 6.54 /
8.49        & -1.26 / -2.2 & -0.6 / -0.98 \\
        \textit{(e)} &  & large 
        & 16.13 / 12.59 & 8.62 / 5.89
        & -0.92 / -1.33 & 1.01 / -0.28
        & 13.94 / 10.97 & 11.08 / 8.62
        & -1.28 / -0.32 & -0.81 / 0.7 \\
        \textit{(f)} &  & large ft
        & 1.91 / 2.56 & 12.2 / 10.98
        & 0.69 / -0.07 & 4.79 / 3.14
        & 2.65 / 2.57 & 7.24 / 7.1
        & 0.04 / 0.07 & 2.68 / 3.26 \\
        
    \bottomrule
    \end{tabular}
    \end{adjustbox}
    \label{table:turn_taking}
\end{table*}
\subsection{Discrete Units Encoder}
The original dGSLM utilized HuBERT \cite{hsu2021hubert} pretrained on the Fisher dataset to obtain better phonetic discriminated representations, followed by training a k-means model to cluster its hidden representation. However, pretraining HuBERT on target domain data is time-consuming and its generalizability is uncertain. To investigate the suitability of existing publicly available speech foundation models for spoken dialogue language model, we explored the base and large versions of HuBERT and WavLM \cite{chen2022wavlm}, which are state-of-the-art speech foundation models.

Additionally, we incorporated HuBERT large fine-tuned (HuBERT large ft) as the speech encoder. HuBERT large ft is obtained by fine-tuning HuBERT large by connectionist temporal classification (CTC) objective. ASR models tend to provide more phonetic information, which might be beneficial for the spoken dialogue language modeling task\footnote{We are of the view that the investigation of speech encoders can inclusively incorporate those trained on labeled data, provided they can effectively model the nuances of non-verbal speech information.}. 
% This diversity in speech encoders allows us to compare their performance and evaluate their effectiveness in the dGSLM framework.

% \subsection{Turn-taking Metrics}

% We adopted the turn-taking definitions from dGSLM \cite{nguyen2023generative}, defining Inter-Pausal Units (IPU) for continuous active time intervals within each channel. Gaps represent silences between different speakers, while pauses occur within the speech of the same speaker. Overlapping speech occurs when two speakers talk simultaneously.

% To evaluate the capability of generating non-verbal vocalization, we define more turn-taking metrics for laughter, which dGSLM only counts without considering its duration\footnote{We noticed that sometimes generated non-stopping laughter, while occurrence fail to catch such mistake}. To address this, we incorporated the duration of laughter. Furthermore, in natural dialogues, one speaker's laughter often prompts the other to laugh shortly after. Thus, we defined Concurrent-Laughter-Rate (CLR) as the occurrence of both speakers laughing within 1 second, capturing the synchronicity of laughter and providing insights into the naturalness of the dialogue flow.

\section{Experiment}
\subsection{Data Setup}
In addition to Fisher Dataset \cite{cieri2004fisher} used in dGSLM \cite{nguyen2023generative}, we scrape about 20k hours of podcast raw audio from Apple Podcast\footnote{\url{https://www.apple.com/apple-podcasts/}}, and after applying our pipeline described in Sec. \ref{sec:pipeline}, we create about 15.6k hours of pseudo-stereo spoken dialogue data. 

To our knowledge, the existing large-scale datasets containing spoken dialogue are GigaSpeech \cite{chen2021gigaspeech} and Spotify Podcast Dataset \cite{clifton2020spotify}. However, GigaSpeech contains at most 7500 hours of raw dialogue data from Podcast and YouTube, which does not increase the dataset too much (after filtering). While Spotify Podcast Dataset comprises about 200k hours of podcast data, Spotify ceased sharing this corpus, prompting us to scrape the podcast data by ourselves.

For validation and evaluation purposes, we randomly chose 1 conversation for each speaker pair in the Fisher, respectively, and we randomly sampled 1\% of the pseudo-stereo data from uniform channels for validation and evaluation. This subset represents a balanced selection for a fair comparison between models.

\subsection{Pseudo-Stereo Data Pipeline}
Our approach to creating pseudo-stereo data involved several steps, utilizing a combination of tools and models for effective speaker diarization, source separation, and speaker verification.

 For speaker diarization, we adopted the pyannote toolkit\footnote{\label{fn:pyannote}\url{https://github.com/pyannote/pyannote-audio}}, we set the size of segments from 30 seconds to 120 seconds depending on how long the two speakers are speaking.

To separate overlapping speech segments and disentangle the speakers, we trained a Sepformer model \cite{subakan2021attention} on the 16k max split of the Libri2Mix dataset \cite{cosentino2020librimix} using the speechbrain toolkit \cite{ravanelli2021speechbrain}. Sepformer model is a SOTA transformer-based model for source separation tasks, providing high-quality separation of overlapping speech.

For speaker verification, we utilized an open-source ECAPA-TDNN checkpoint\footnote{\url{https://huggingface.co/speechbrain/spkrec-ecapa-voxceleb}} \cite{desplanques2020ecapa, ravanelli2021speechbrain}. ECAPA-TDNN model is a deep neural network architecture designed for speaker recognition tasks, offering robust speaker embedding capabilities.

\subsection{Model Training}
For speech encoder, we select official checkpoints of HuBERT \cite{hsu2021hubert} base/large/large ft, and WavLM \cite{chen2022wavlm} base+ /large. For each types of speech encoder, we use two types of training data to train the dialogue language model: the Fisher dataset \cite{cieri2004fisher} and the Fisher dataset combined with pseudo-stereo data. This approach allows us to investigate the impact of incorporating pseudo-stereo data on the performance of the spoken dialogue modeling.

We follow the same architecture and training procedure of dialogue language model and vocoder in dGSLM \cite{nguyen2023generative} but with different dataset and speech encoders. We selected 100 hours of training data to clustered the last layer of representations from speech encoder to encode speech into 500 clusters.

For vocoder, we utilized single-channel data from Fisher\footnote{Generated by the script: \url{https://gitlab.nrp-nautilus.io/ar-noc/nemo/-/blob/master/scripts/process_fisher_data.py}} combined with VCTK \cite{yamagishi2019vctk} for better generalizability of vocoder across Fisher and Podcast data.

\subsection{Evaluation Metrics}
\subsubsection{Turn-taking Event Statistics}

\begin{table}[h]
    \caption{The human evaluation of meaningfulness, with 95\% confidence intervals. Semantic coherence generally improves with the addition of stereo audio, as indicated by bold scores.}
    \centering
    \vspace{1pt}

    \begin{tabular} {l c c|@{\extracolsep{4pt}}cc@{}}
        \toprule
         & \multirow{2}{*}{Encoder} & \multirow{2}{*}{Type} & \multicolumn{2}{c}{M-MOS} \\
        \cline{4-5}
         &  &  & Fisher & Podcast \\
        \midrule
        \textit{(a)} & Ground Truth & 
        & 3.55 ± 0.23 & 4.2 ± 0.2 \\
        \midrule
        \multicolumn{5}{l}{\textbf{\textsc{Fisher}}} \\
        \textit{(b)} & \multirow{2}{*}{WavLM} & base+
        & 3.14 ± 0.23 & 3.31 ± 0.24 \\
        \textit{(c)} &  & large
        & 3.45 ± 0.25 & 3.17 ± 0.29 \\
        \midrule
        \textit{(d)} & \multirow{2}{*}{HuBERT} & base
        & 3.29 ± 0.27 & 3.36 ± 0.31 \\
        \textit{(e)} &  & large ft
        & 3.50 ± 0.31 & 3.67 ± 0.30 \\
        \midrule
        \multicolumn{5}{l}{\textbf{\textsc{Fisher + Pseudo-stereo Aduio}}} \\
        \textit{(f)} & \multirow{2}{*}{WavLM} & base+
        & \textbf{3.25 ± 0.24} & 3.18 ± 0.27 \\
        \textit{(g)} &  & large
        & 3.33 ± 0.37 & \textbf{3.49 ± 0.24} \\
        \midrule
        \textit{(h)} & \multirow{2}{*}{HuBERT} & base
        & \textbf{3.47 ± 0.25} & \textbf{3.43 ± 0.17} \\
        \textit{(i)} &  & large ft
        & \textbf{3.65 ± 0.25} & \textbf{3.74 ± 0.17} \\
    \bottomrule
    \end{tabular}

    \label{table:mos}
\end{table}
\label{sec:turn_taking_metric}
We follows the definition of turn-taking events in \cite{nguyen2023generative}. Inter-Pausal Unit (IPU) indicates the frames that a speaker is speaking; Gap is the silence frames occur between two speakers; Pause is the silence frames occur within the same speaker; Overlap indicates the frames that two speakers are both speaking.

We generated samples by promting the dialogue language model by 30 seconds dialogue from test set, and hypothesize 60 seconds continuation. The decoding temperature is set to 1.0, and sampling among the top 20 possible units. We find silence frames by applying VAD\footref{fn:pyannote} to each channel separately, and compute IPU, Gap, Pause, and Overlap. 
% We compute IPU, gap, pause, overlap by applying VAD\footref{fn:pyannote} to each channels separately. Noting that Fisher and pseudo-stereo data share similar trend on IPU, gap, pause, and overlap, so we did report the individual results for these metrics.

% , and apply ASR \footnote{whisper medium.en} to each channels separately, and compute the number of words over time to get WPS.

% While WPS varies a lot in the two dataset, so we report their results separately.

% \input{tabs/wps}

\subsubsection{Human Evaluation}
We conducted human evaluation to assess semantic coherence of our dialogue generation models.

We chose 30 10-second prompts each from the test sets of the Fisher dataset and the pseudo-stereo data. These prompts were carefully selected for high quality, considering factors like audio clarity and minimal noise. Then using the same sampling configuration as described in Section \ref{sec:turn_taking_metric}, we generated 20-second continuations for each prompt. Workers are asked to rate the samples based solely on the content and semantic coherence from 1 to 5.

We employed MTurk\footnote{\url{https://www.mturk.com/}} to recruit workers. Every hit is composed by 20 samples, and each sample is evaluated by 3 workers. We rejected all workers that their work time are less than 10 minutes, or gives 1 or 2 for ground truth.
\section{Results}

\subsection{Turn-taking Behaviors}
% \subsubsection{Turn-taking Event Statistics}
Table \ref{table:turn_taking} presents the turn-taking events discussed in Section \ref{sec:turn_taking_metric} for different settings. We report their differences from the ground truth, with a slash separating whether the pseudo-stereo data was used during training.

Our initial observations reveal that most cases (row (b), (c), (d), (f)) successfully capture turn-taking behaviours similar to real dialogue. Combining with pseudo-stereo data does not further improve the performance, which indicate that limited dialogue data are suficient to model these turn-taking behavior.

However, HuBERT large (row (e)) fail to correctly model spoken dialogue. This discrepancy primarily stems from the vocoder trained on units encoded by HuBERT large, which struggles to accurately resynthesize audio (even prompt can not be correctly reconstruct, see our demo page\footref{fn:demo}). It suggest that the representations in the last layer of HuBERT large contains little information about original audio. Conversely, fine-tuning the model with ASR gives better phoneme discrimination for representations in the last layer, making it easier to reconstruct the original speech from the discrete units. This makes the ASR fine-tuned model more suitable for our case

We do not report the performance of HuBERT large in the following results, for it is too corrupted.

% \subsubsection{Speak Speed}
% Table \ref{table:wps} shows the words-per-second (WPS) difference with the ground truth. Models tend to speak quicker than the ground truth, 

\subsection{Human Evaluations}
Table \ref{table:mos} reports the M-MOS of generated speech from prompts in Fisher and Podcast respectively. Combining pseudo-stereo data during training improves the semantic coherence within speech as seen in row (h), (i), (f) for Fisher, (g) for Podcast.

Additionally, HuBERT large ft model combined with pseudo-stereo data performs exceptionally well among prompts from both Fisher and Podcast. In fact, the performances even surpasses the ground truth of Fisher dataset.

This improvement may stem from the nature of the datasets. Fisher, being a telephonic dialogue dataset, tends to have more casual conversations with less information density. On the other hand, podcasts often have a central theme, resulting in more semantically coherent dialogue. The large scale of the pseudo-stereo data, when compared to Fisher, might also influence the models to generate dialogue resembling podcasts, thus enhancing its meaningfulness.

It's worth noting that we  did not directly report the naturalness of generated audio in Table \ref{table:mos}, although we conducted an Naturalness MOS (N-MOS) survey. We evaluate the same data as M-MOS, along with the resynthesized ground truth. The maximum difference between model generated continuation and resynthesized ground truth was no greater than 0.4; while ground truth resynthesized by different vocoder can vary by 0.8, 

This observation suggests that our current resynthesis approach is not robust enough and often generates noise or distorted speech, leading to testers assessing mainly the vocoder's performance rather than the naturalness of turn-taking behavior.
\section{Conclusion}
This work presents a novel solution to address the scarcity of stereo dialogue data in Spoken Dialogue Modeling. Our developed pipeline transforms single-channel dialogue into pseudo-stereo data, significantly expanding our training dataset from 2,000 to 17,600 hours, and we open source our dataset for future research.

Additionally, we explored the effectiveness of using discrete units from various speech foundation models for dialogue generation. Our experiments revealed that employing HuBERT large ft as our speech encoder, and combined with pseudo-stereo data as training data notably improved the semantic coherence of the generated dialogue.

In conclusion, our contributions provide practical solutions to the challenges associated with limited stereo dialogue data and speech encoder selection. The integration of pseudo-stereo data and HuBERT large ft encoder has led to substantial improvements in spoken dialogue synthesis. However, it's important to note that the current unit-to-speech vocoder is not yet robust enough, impacting the quality of the generated audio. This suggests a potential avenue for future research to further enhance the audio quality of speech resynthesis from discrete units.

\bibliographystyle{IEEEtran}
\bibliography{mybib}

\end{document}